\documentclass{article}

\usepackage[preprint]{neurips}

\usepackage[utf8]{inputenc} 
\usepackage[T1]{fontenc}    
\usepackage{url}            
\usepackage{booktabs}       
\usepackage{amsfonts}       
\usepackage{nicefrac}       
\usepackage{microtype}      
\usepackage{xcolor}         

\usepackage{amsmath}
\usepackage{amssymb}
\usepackage{mathtools}
\usepackage{amsthm}

\usepackage{multirow}
\usepackage{graphicx}
\usepackage{wrapfig}
\usepackage{caption}
\usepackage{pifont}

\newcommand{\cmark}{\ding{51}} 

\usepackage[hidelinks,breaklinks=true,colorlinks,citecolor=citecolor,linkcolor=linkcolor]{hyperref}
\usepackage[capitalize,noabbrev]{cleveref}

\usepackage{colortbl}

\usepackage{array}    
\usepackage{tocloft}
\usepackage {titletoc}
\definecolor{iccvblue}{rgb}{0.21,0.49,0.74}

\definecolor{citecolor}{HTML}{2980b9}
\definecolor{linkcolor}{HTML}{c0392b}
\definecolor{sem}{HTML}{2E75B6}
\definecolor{tok}{HTML}{F3B000}

\usepackage{subcaption} 

\title{LongDWM: Cross-Granularity Distillation for Building a Long-Term Driving World Model}

\author{Xiaodong Wang$^{1,2}$  \quad Zhirong Wu$^{1}$ \quad Peixi Peng$^{1,2}$\thanks{Corresponding author.}   \\
 {$^{1}$Peking University \quad $^{2}$Peng Cheng Laboratory} \\
{\tt\small\{wangxiaodong21s@stu., pxpeng\}@pku.edu.cn}
}

\begin{document}

\maketitle

\vspace{-5mm}
\begin{abstract}
Driving world models are used to simulate futures by video generation based on the condition of the current state and actions.  However, current models often suffer serious error accumulations when predicting the long-term future, which limits the practical application. Recent studies utilize the Diffusion Transformer (DiT) as the backbone of driving world models to improve learning flexibility. However, these models are always trained on short video clips (high fps and short duration), and multiple roll-out generations struggle to produce consistent and reasonable long videos due to the training-inference gap. To this end, we propose several solutions to build a simple yet effective long-term driving world model. First, we hierarchically decouple world model learning into large motion learning and bidirectional continuous motion learning. Then, considering the continuity of driving scenes, we propose a simple distillation method where fine-grained video flows are self-supervised signals for coarse-grained flows. The distillation is designed to improve the coherence of infinite video generation. The coarse-grained and fine-grained modules are coordinated to generate long-term and temporally coherent videos. In the public benchmark NuScenes, compared with the state-of-the-art front-view model, our model improves FVD by 27\% and reduces inference time by 85\% for the video task of generating 110+ frames. More videos (including 90s duration) are available at \url{https://Wang-Xiaodong1899.github.io/longdwm/}.

\end{abstract}
\section{Introduction}

World models are used to predict the environment dynamics of different actions based on the current state~\cite{ha2018world,lecun2022worldmodel}, which is very important for autonomous driving. Earlier works design world models in latent feature space~\cite{hafner2019dreamer,hafner2020dreamerv2,hafner2023dreamerv3}, and could facilitate the learning of control policies~\cite{ebert2018visual,dosovitskiy2017carla,tassa2018dmc}. To improve the interpretability and interoperability to the human driver,  several works 
use controllable video generation technology to build the driving world model and predict the future as video, which has made great progress in the past few years~\cite{hu2023gaia,wang2023drivedreamer,wang2023driveWM,gao2024vista}, and video generators also achieved promising results as world simulators~\cite{openai2023sora,kong2024hunyuanvideo}. For driving world models, it is crucial to have good long-term prediction capabilities in order to provide more reasonable and accurate guidance for current state decisions.

\begin{figure*}
    \centering
    \includegraphics[width=0.95\linewidth]{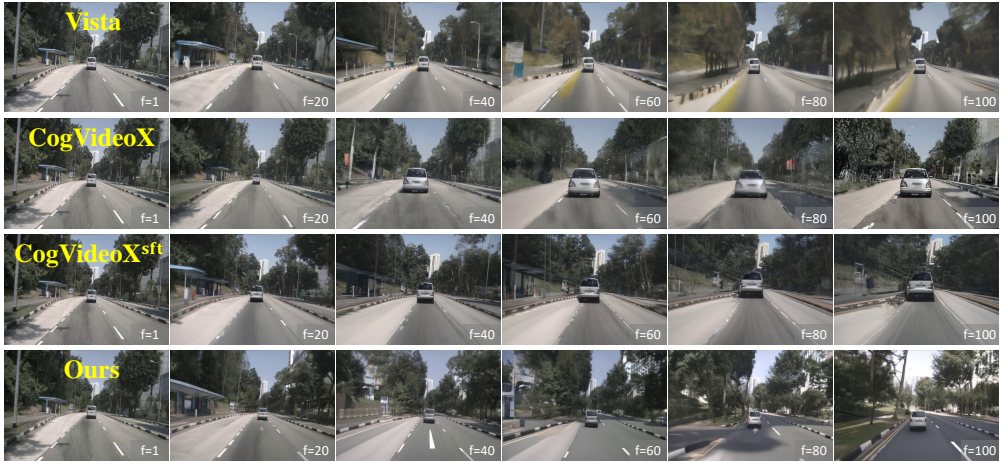}
    \caption{\textbf{Long video generation comparison in the autonomous driving scenario.} The results include Vista~\cite{gao2024vista}, CogVideoX~\cite{yang2024cogvideox}, CogVideoX$^\texttt{sft}$ and ours, the input first frame (f=1) is from the validation set of NuScenes~\cite{caesar2020nuscenes}. The result of Vista shows extremely blurred frames (e.g., yellow stripe), and the results of CogVideoX and variant show unrealistic motion, while our model generates a realistic long-term future.}
    \vspace{-10pt}
    \label{fig:teaser}
\end{figure*}

However, serious error accumulations are easy to observe when driving world models predict the long-term future, which limits the practical application. The essential reason is that long video generation is a challenging task, and needs to bridge the gap between general scenarios and driving scenarios, especially in driving tasks with large motion. It requires generated videos to have long-term coherent, reasonable and accurate scene development. The first challenge comes from the training-inference gap. Current driving world models utilize diffusion models, and they train their model in a short clip with high fps, such as Vista~\cite{gao2024vista}, a state-of-the-art model based on a U-Net backbone trained on 25-frame clips with 10 fps, which encounters serious error accumulation when rolling out long videos, as shown in the first row in Fig.~\ref{fig:teaser}. Training with short clips and continuous frames can be regarded as a ``free lunch'' approach, since frames with short duration and high fps with smoother temporal distribution for easier learning, and it is widely used in driving or the general domain, such as a state-of-the-art model CogVideoX~\cite{yang2024cogvideox} based the Diffusion Transformer (DiT) backbone~\cite{peebles2023DiT}. But this ``free lunch'' approach causes the training-inference gap. That is, longer videos often contain large transitions of scenes, while these models could only predict smoother temporal motion, resulting in serious error accumulation.

The second challenge is the gap between general scenarios and driving scenarios. If we use a general video generator such as CogVideoX to roll out long videos from a static driving scene, as shown in the second row in Fig.~\ref{fig:teaser}, although it is a powerful model utilizing DiT in open-domain, and alleviates the accumulation of edge distortion to a some extent, it still produce blurred frames and unrealistic motion. The potential reason is that general scenarios and driving scenarios differ significantly in some aspects, such as motion, scene development, and so on. A direct way to bridge the gap is finetuning in the driving scenario, but if using ``free lunch'' approach like CogVideoX$^\texttt{sft}$~\cite{yang2024cogvideox}, the unrealistic motion still exists, such as "The street scene is changing, but the car in front remains stationary." as shown in the third row in Fig.~\ref{fig:teaser}. These two challenges have not yet been addressed for driving world models. This study attempts to address these challenges and builds a simple long-term driving world model, and our model's long video prediction is shown in the last row in Fig.~\ref{fig:teaser}. Our prediction result is significantly better than others, showing clear details, consistent temporal dynamics, and realistic driving scene development.


To address the above challenges, we propose several solutions to build a simple yet effective long-term driving world model. For failure patterns of training-inference gap, previous works~\cite{zhao2025riflex} point out that the temporal repetition, blurred details, and unrealistic motion exist when rolling out long videos using CogVideoX and HunyuanVideo~\cite{kong2024hunyuanvideo}. This phenomenon also exists in the predictions of driving world models, such as the error accumulations in the results of Vista and CogvideoX$^\texttt{sft}$. To eliminate the gap, some related works~\cite{zhao2024moviedreamer,yin2023nuwaxl} use textual scripts to generate keyframes for a long video and use image-to-video models to extend keyframes. But these works only generate movies or cartoons. To address the ``free lunch'' problem for driving world models, we decouple the long-term learning into large motion learning (e.g., scene transitions) and small continuous motion learning (e.g., car motions). Decoupled learning naturally eliminates the gap, because world models can first predict large motions and then fill in with small continuous motions when making long-term predictions. 

For the gap between general scenarios and driving scenarios, previous works~\cite{zhao2024moviedreamer,yin2023nuwaxl} only focus on text-to-video, emphasizing that the plot development of long videos needs to conform to given scripts. While in driving world models, the target is to predict the long-term future from the current state, i.e., the current image, and the results of large motion prediction should maintain coherence with the current state. To achieve this, for large motion learning, we construct a Coarse DiT training on coarse frames for causal prediction, and video tokens are encoded independently from each video frame using a low fps. No temporal compression is used to preserve the details of each frame. For small continuous motion learning, a Fine DiT is trained on temporally compressed video tokens for causal and bidirectional predictions, where the video frames have a high fps and adjacent frames are compressed. Meanwhile, considering the continuity characteristics of driving scenes, such as ``continuous street scene changes gradually lead to the change from one scene to a new scene'', simply predicting large motions leads to scene mutations and distortion. To this end, we propose a simple flow distillation method. Intuitively, fine-grained video tokens have better consistency than coarse-grained video tokens, so we use the fine-grained priors to guide the coarse-grained prediction. Given well-trained Fine DiT and Coarse DiT, we first sample a continuous frame sequence where coarse frames are marked. We input all coarse frames into a trainable Coarse DiT to obtain the coarse flow, and input all continuous frame segments where the first and last are coarse frames into the frozen Fine DiT to obtain fine flows, where the flow is a one-step denoising output. The distillation loss is an L2 loss between coarse flows and fine flows at corresponding positions. This distillation regards fine flows as self-supervised signals, thereby encouraging the Coarse DiT to make more consistent predictions across frames. Additionally, we propose a novel warp-guided controllable video prediction method for Fine DiT to improve the controllability.
Overall, our contributions are three-fold:
\begin{itemize}
\item We propose a simple yet effective long-term driving world model, where the world model learning is decoupled into hierarchical learning for the first time, including large motion learning and bidirectional continuous motion learning.

\item We propose a flow distillation method that fine-grained video flows are self-supervised signals for coarse-grained video flows, prompting the consistency of coarse token predictions.

\item Experiments on NuScenes dataset demonstrate our model achieves state-of-the-art performances on the FVD metric in all settings. In particular, compared with the state-of-the-art front-view model Vista, our model improves FVD by 27\% and reduces inference time by 85\% for the video task of generating 110+ frames.
\end{itemize}

\section{Related Work}

\subsection{Video Generation}
\paragraph{Video Diffusion Models}
Diffusion models have made great progress in video generation. Early works usually leverage a pretrained text-to-image model~\cite{rombach2022sdm} and insert temporal layers into the base architecture and continue train on video-text paired data~\cite{yin2023nuwaxl,blattmann2023align,guo2023animatediff,blattmann2023svd}.~\cite{blattmann2023align} inserts temporal convolution and temporal layers into base diffusion U-Net to adapt the video generation task.~\cite{yin2023nuwaxl} also inserts various temporal layers and other conditional layers into the base architecture. Some works only train extra temporal layers~\cite{blattmann2023align,guo2023animatediff} and some works fine-tune the full models~\cite{blattmann2023svd}.

Recently, diffusion transformer (DiT) models have shown great improvement in video quality. ~\cite{peebles2023DiT} utilizes Transformer as the backbone of diffusion models, which prompts the text-to-video to reach a new milestone such as Sora~\cite{openai2023sora}. Following Sora, there are some impressive open-sourced models, such as Vidu~\cite{bao2024vidu}, CogVideoX~\cite{yang2024cogvideox}, HunyuanVideo~\cite{kong2024hunyuanvideo}, etc. These models try to bridge the gap between the open-source models and closed-source models (e.g., Sora), and have already surpassed Sora in some aspects. Overall, a lot of work in high-quality data curation, advanced architecture design, and infrastructure improvements has facilitated the development of video diffusion models.

\paragraph{Long Video Generation}
The main challenge for this task is that long videos encounter error accumulation, resulting in blurring and distortion of videos. Naturally, auto-regressive models are more suitable for this since they can receive variable video context and generate video frames with variable length, which can alleviate error accumulation using slide windows~\cite{liang2022nuwainfinity,henschel2024streamingt2v}, but also face high memory pressure for a long sequence of video tokens. Diffusion models are always trained on video frames with a fixed length, and most works train their models on clips with high fps and short duration, due to GPU memory limitation~\cite{yang2024cogvideox,kong2024hunyuanvideo,bao2024vidu}.  Some works utilize hierarchical approaches that first generate keyframes and then interpolate continuous frames between them~\cite{yin2023nuwaxl,zhao2024moviedreamer}. However, these works only focus on text-to-video, emphasizing that the plot development of long videos needs to conform to given scripts. While for driving world models, the target is that future predictions should maintain coherence with the current state. We propose a novel distillation method to guide the long video generation, instead of treating different granularities independently as in previous works.


\subsection{Driving World Model}
Driving world models leverage the world model to predict the environment dynamics of different driving actions based on the current state. As predictive and generative models have achieved great progress, based on these, driving world models have better instruction-following capabilities and higher-quality predictions. GAIA-1~\cite{hu2023gaia} proposes a driving world model based on diffusion models that leverages video, text, and action inputs to generate future scenarios. Some works~\cite{wang2023drivedreamer,jia2023adriver,wang2025prophetdwm} build multi-modal driving world models to support video generation and action prediction. Besides video and text, some works~\cite{wang2023driveWM,gao2023magicdrive} utilize 3D annotations or multi-view inputs to predict future scenarios with nuanced 3D geometry. Recent works expand driving world models to the evolution predictions of 3D occupancy~\cite{zheng2025occworld,xu2025temporal} or holistic models for perception, prediction and planning~\cite{zheng2024doe}. Vista~\cite{gao2024vista} proposes a driving world model with higher resolution and versatile controllability that can generalize to diverse scenarios. The most related works~\cite{wu2024holodrive,gao2024magicdrivedit,jiang2024dive,guo2025dist,li2024uniscene} focus on multi-view generation or occupancy prediction using DiT, but train with short clips. However, these models face error accumulations when predicting the long future, while our method systematically addresses this.

\begin{figure*}
    \centering
    \includegraphics[width=\linewidth]{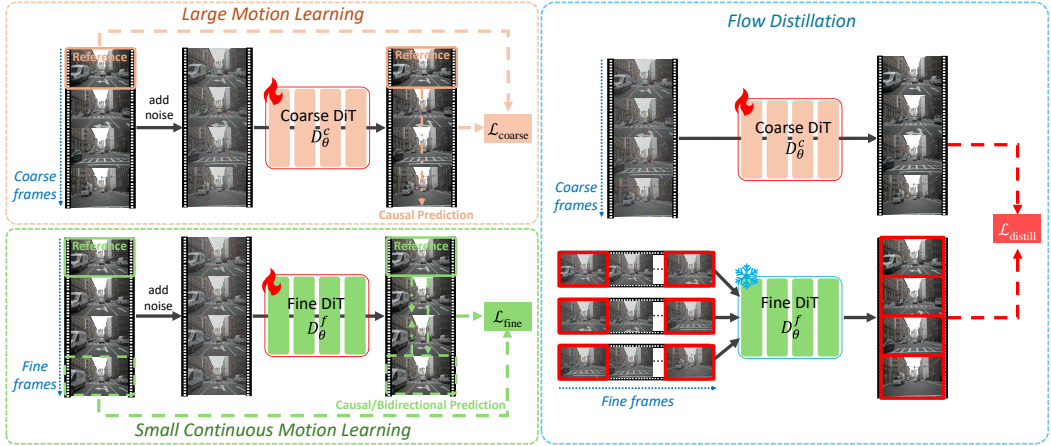}
    \caption{\textbf{The overall distillation framework.} First, we decouple the long-term world model learning into a large motion learning and a small continuous motion learning by designing a Coarse DiT and a Fine DiT to adapt to different granularities. Then, we propose a novel flow distillation method between different granularities, i.e., using fine flows' better priors to distill coarse flows, which prompts the Coarse DiT to produce more consistent predictions. After a few hundred steps, our distillation process effectively tunes the Coarse DiT well.}
    \vspace{-10pt}
    \label{fig:main}
\end{figure*}

\section{Methodology}

\subsection{Coarse Diffusion Transformer}
\label{sec:CDiT}
Previous methods train diffusion U-Nets or Transformers on video clips with short duration and high fps, since these types of clips have smoother temporal distribution for learning, but these models are faced with error accumulation after multiple rollouts for long video generation~\cite{gao2024vista}. 
To avoid relying on the above ``free lunch" approach, this paper proposes a novel Coarse Diffusion Transformer (CDiT) trained on a few frames from long-duration clips. The CDiT model offers more robust forward-looking predictions and is capable of predicting large dynamic information.

Given a long video clip $v$, we first sampling $K$ frames using a low fps (such as fps=1), denoted as coarse video frames $v^c=\{v^{c1}, v^{c2}, \cdots, v^{cK}\}$. There is greater dynamic information between these coarse-grained frames, which the CDiT model needs to learn. These coarse video frames are no longer suitable for causal encoding by 3D-VAE, so we use 3D-VAE to encode each coarse frame independently. The coarse video latents are denoted as $x^c=\{x^{c1}, x^{c2}, \cdots, x^{cK}\}$. Given the latent $x_0$ draw from data distribution $q(x_0)$, and $x_1,\cdots, x_T$ are latents of the same dimensionality as $x_0$, the diffusion process is defined to a Markov chain as below:
\begin{equation}
    q(x_t|x_{t-1}) = \mathcal{N}(x_t;\sqrt{\alpha_t}x_{t-1},(1-\alpha_t)\mathbf{I}),
\end{equation}
where $\alpha_t>0$ is scalar according to a specific noise scheduler at timestep $t$.
We regard $x^c_0:=x^c$ as the start point of the forward process, and add random Gaussian noise $\epsilon$ to the initial sample:
\begin{equation}
    x^c_t = \sqrt{\bar{\alpha}_t}x^c_0 + (1-\bar{\alpha}_t)\epsilon,\  \epsilon \sim \mathcal{N}(\mathbf{0},\mathbf{I}),
\end{equation}
where $t$ is a random timestep and $\bar{\alpha}_t:=\Pi^t_{s=1}\alpha_s$. 

The large dynamic information requires more fine-grained video descriptions. Unlike Vista~\cite{gao2024vista} that uses overly simplistic video annotations, we utilize a multi-modal large language model (MLLM)~\cite{zhang2024llavanextvideo} to annotate videos with more detailed annotations. More details can be found in the Appendix.
Given a detailed video prompt, we utilize T5~\cite{raffel2020T5} to encode the prompt to a text embedding $p$. Then, we design a image-to-video prediction task in the latent space as follows:
\begin{equation}
    \mathcal{L}_\text{diffusion} = \mathbb{E}_{x^c, p, \epsilon, t} \| D^c_{\theta}(x^c_t, t, p, x^{c1})-x^c \|^2_2,
\end{equation}
where $D^c_\theta$ is the CDiT denoiser that shares the same architecture with CogVideoX~\cite{yang2024cogvideox}, and $x^{c1}$ denotes the first coarse frame of a random sampling clip, and the embedded way is the same as~\cite{wang2023learning}.

Learning the large dynamic information would affect the structure and details of each individual frame, so we introduce a latent structure preservation loss as follows:
\begin{equation}
    \mathcal{L}_\text{struct} = \mathbb{E}_{x^c, p, \epsilon, t} \| \mathcal{H}(D_{\theta}(x^c_t, t, p, x^{c1})) - \mathcal{H}(x^c) \|^2_2,
\end{equation}
where $\mathcal{H}$ denotes the 2D high-pass filter of frequency domain in latent space. The final learning objective of CDiT is defined as below:
\begin{equation}
    \mathcal{L}_\text{coarse} = \mathcal{L}_\text{diffusion} + \beta_s\mathcal{L}_\text{struct},
\end{equation}
where $\beta_s$ controls the importance of the structure preservation.

\subsection{Fine Diffusion Transformer}
\label{sec:FDiT}
Besides the ability to predict large dynamic information, fine-grained details and reasoning information should also be included in the long video prediction process. For example, if there is a car around the ego car in the previous keyframe, but the car is gone in the current keyframe, the correct reasoning should be to fill in a segment of the car slowly driving away between the two frames. To learn the video reasoning ability, we design a versatile Fine DiT (FDiT) that learns both video prediction and video interpolation. In this stage, we sample $K$ continuous frames from short-duration video clips, denoted as $v^f=\{v^{f1}, v^{f2}, \cdots, v^{fK}\}$. The continuous frames are encoded to latents $x^f=\{x^{f1}, x^{f2}, \cdots, x^{fK}\}$ by a 3D-VAE. We reuse the MLLM's annotations and design the training objective as below:
\begin{equation}
    \mathcal{L}_\text{fine} = \mathbb{E}_{x^f, p, \epsilon, t} \| D^f_{\theta}(x^f_t, t, p, x^{f1}, \ddot{x}^{fK})-x^f \|^2_2,
\end{equation}
where $D^f_{\theta}$ denotes the FDiT denoiser, $\ddot{x}^{fK}$ denotes the last frame $x^{fK}$ that would be dropped with a fixed ratio, and FDiT learns causal prediction and bidirectional prediction, that is, video prediction and video interpolation, and the two tasks can help each other.

\subsection{Fine-flow $\rightarrow$ Coarse-flow Distillation}
The CDiT would scarify consistency to achieve large dynamic prediction, unlike previous hierarchical U-Net models neglecting this problem~\cite{yin2023nuwaxl}, we instead propose a novel distillation method that distills prior knowledge from fine flow into the coarse flow of CDiT. Previous methods utilize distillation to reduce denoising steps or computing time in the same granularity flows, i.e., short-duration videos~\cite{yin2024DMD,yin2024DMD2,yin2024slow}. We propose a simple distillation method between different granularity. The fine flows in FDiT can give more guidance to coarse flows since FDiT is good at predicting fine-grained smooth temporal changes.

The overall illustration is shown in Fig~\ref{fig:main}. After training both CDiT and FDiT models, we can initialize the distillation process. Given a video clip $v$, we first sample a sequence of frames using a high fps same as Sec.~\ref{sec:FDiT}, and coarse frames are marked, the sequence is denoted as below:
\begin{equation}
    v_d=\{\underbrace{v^{c1}, v^{f2},\cdots,v^{c2}}_{K\  \text{items}}, \cdots,v^{cK}\},
\end{equation}
where $v_d$ can be divided into segments of $K-1$ consecutive frames, and the first and last frames of each segment are coarse frames. To distill the CDiT to have more consistent predictions, we utilize the frozen FDiT to separately cope with latents of each segment, and then the first and last latent of each segment are concatenated together. Specifically, we first randomly sample a timestep $t$ and a Gaussian noise with the same shape of latents. Then, each segment share the same timestep and noise to predict the latents using one-step denoising. The coarse frames $\{v^{c1},v^{c2},\cdots,v^{cK} \}$ are also added by same noise with timestep $t$ and predict the coarse latents. The distillation loss is defined as below:
\begin{equation}    
    \mathcal{L}_\text{distill} = \| D^f_{\theta, \text{frozen}}(v_{(d,t)})[c1, c2, \cdots, cK] - D^c_{\theta}(v_{(d,t)}[c1, c2, \cdots, cK]) \|^2
\end{equation}

where the teacher model is a frozen FDiT well-trained from Sec.~\ref{sec:FDiT} and the student model is initialized from a CDiT well-trained from Sec.~\ref{sec:CDiT}. Our distillation requires only a few hundred steps to tune the student model well.

\subsection{Warp-guided Controllable Video Prediction}
Previous methods primarily adapt camera trajectory controllable video generation by fusing trajectory or camera pose features into models, but often perform poorly for complex trajectories~\cite{he2024cameractrl,bahmani2024vd3d,gao2024vista}. Recently, some studies~\cite{hou2024training,bian2025gs} improve the generation by incorporating 3D priors. 
\begin{wrapfigure}{r}{0.5\linewidth}
    \centering
    \vspace{-10pt} 
    \label{fig:traj_method}
    \includegraphics[width=\linewidth]{figs/traj-method.pdf}
    \caption{\textbf{Warp-guided trajectory controllable video training.} Given an input trajectory, we first transform it into camera poses and utilize a depth-free image warping to obtain warped subsequent frames. These frames are fed into FDiT along with a simple prompt to predict the clean frames.}
    \vspace{-20pt} 
\end{wrapfigure}
Inspired by this, we propose a novel trajectory control method that leverages 3D information without the need for reconstruction or additional annotations. Given an input trajectory, Vista~\cite{gao2024vista} only inputs it into diffusion models as an extra condition, while in our process, we leverage trajectory priors to obtain warped future frames to stabilize the control ability. More details can be found in Appendix~\ref {app:detal_warp}.

\paragraph{Video prediction with warped images.} As shown in Fig.~3, our model is built on our Fine DiT and enhances the trajectory controllable video prediction by incorporating warped subsequent frames for guidance. However, the warped frames have some distortions, such as inconsistent rotation and varying motion speeds. These issues stem primarily from our simplistic assumption of uniform depth. To address this, after the warped images are encoded by the 3D-VAE, we first pass them through a trainable patch embedding layer before injecting them into each block of the DiT. Warped features prompt our model to predict more accurate controllable predictions.

\begin{table*}
\caption{\textbf{Comparison of prediction fidelity on NuScenes validation set.} 
Our model outperforms the state-of-the-art driving world models. $^\dagger$ denotes our evaluation using open source checkpoints from~\cite{gao2024vista}. \textcolor{gray}{Reconstruction} denotes using VAE from CogVideoX~\cite{yang2024cogvideox}.}
\label{tab:fvd}
\centering
    \begin{tabular}{l|cccc}
    \toprule
    \textbf{Model}   & \textbf{Extra data} & \textbf{Backbone}  & \textbf{FID↓} & \textbf{FVD↓} \\
    \midrule
    DriveGAN~\cite{kim2021drivegan} & $\times$  & GAN    & 73.4  & 502.3 \\
    DriveDreamer~\cite{wang2023drivedreamer} & $\times$ & U-Net &  52.6  & 452.0 \\
    WoVoGen~\cite{lu2025wovogen} & $\times$ & U-Net   &  27.6   &  417.7  \\
    Drive-WM~\cite{wang2023driveWM} & $\times$  & U-Net &  15.8   &   122.7  \\
    GenAD~\cite{yang2024generalized} & $\times$   & U-Net  &  15.4  & 244.0 \\
    GenAD~\cite{yang2024generalized} & $\cmark$   & U-Net  &  15.4  & 184.0 \\
    \textcolor{gray}{Vista~\cite{gao2024vista}}& $\cmark$      &  \textcolor{gray}{U-Net}   & \textcolor{gray}{6.9}   &  \textcolor{gray}{89.4} \\
    Vista~\cite{gao2024vista}$^\dagger$  & $\cmark$     &  U-Net   & \textbf{7.6}   &  128.5 \\
\midrule

    UniMLVG~\cite{chen2024unimlvg} & $\cmark$  & DiT & 30.5  & 149.7 \\

    CogVideoX$^{\texttt{sft}}$~\cite{yang2024cogvideox} & $\times$ & DiT & 15.8  & 117.0 \\

    \midrule
    Ours & $\times$  & DiT       &     12.3      & \textbf{102.9} \\
    \textcolor{gray}{Reconstruction} & $\times$ & \textcolor{gray}{DiT}  &  \textcolor{gray}{4.9}     &  \textcolor{gray}{31.3} \\
    \bottomrule
    \end{tabular}
\end{table*}


\begin{table*}
\caption{\textbf{Long-term video prediction on NuScenes validation subset.} 
We report the FVD and inference time for various durations.}
\label{tab:long}
\centering
\resizebox{\textwidth}{!}{
    \begin{tabular}
    {ll|cccccc}
    \toprule
    \textbf{Duration} & \textbf{Model} & I2VGen-XL & DynamiCrafter& CogVideoX-I2V-5B &   SVD    &   Vista  & Ours  \\
     \midrule
    25f$\sim$2.5\text{s}( &  \textbf{FVD↓} &768.1 & 357.5 & 400.3  & 194.6    & \cellcolor{gray!40} 174.1     &   188.5   \\
        1st rollout)       &  \textbf{Time↓}   & 78s & 88s & 140s &  44s   &   95s     &  \cellcolor{gray!40} 35s \\
\midrule
     69f$\sim$6.9\text{s}( & \textbf{ FVD↓ } & 1214.2  & 602.4 & 482.5 & 352.4    &   277.5   &   \cellcolor{gray!40} 242.2      \\
        3rd rollout)    & \textbf{ Time↓ }  & 234s & 220s & 660s & 132s  &   285s     &   \cellcolor{gray!40} 70s  \\

\midrule
    113f$\sim$11.3 \text{s}( & \textbf{ FVD↓ }&1616.3 & 1155.5 & 533.7  & 753.0     &    398.5   &    \cellcolor{gray!40} 289.5   \\
           5th rollout)      & \textbf{ Time↓ }  & 390s & 352s & 990s &   220s  &  475s     &   \cellcolor{gray!40} 70s  \\
                         
    \bottomrule
    \end{tabular}
}
\vspace{-5pt}
\end{table*}
\section{Experiments}

\paragraph{Implementation Details}
All training and evaluations are based on NuScenes benchmark~\cite{caesar2020nuscenes}. Our models are initialized with CogVideoX-2B~\cite{yang2024cogvideox}. Then our models are trained with a resolution of 720$\times$480 on the training set. We evaluate the video prediction quality on the validation set utilizing metrics FVD and FID. We choose state-of-the-art models Vista~\cite{gao2024vista}, SVD~\cite{blattmann2023svd}, DynamiCrafer~\cite{xing2025dynamicrafter}, I2VGen-XL~\cite{zhang2023i2vgen}, and CogVideoX-I2V-5B~\cite{yang2024cogvideox} as baselines for fair comparison. For driving models using DiT, we compare UniMLVG~\cite{chen2024unimlvg}, since other models~\cite{gao2024magicdrivedit,jiang2024dive,guo2025dist,li2024uniscene} only support multi-view predictions. Please see Appendix~\ref{app:training} for more details.

\subsection{Results of Generation Quality and Fidelity}
\begin{figure*}
    \centering
    \includegraphics[width=\linewidth]{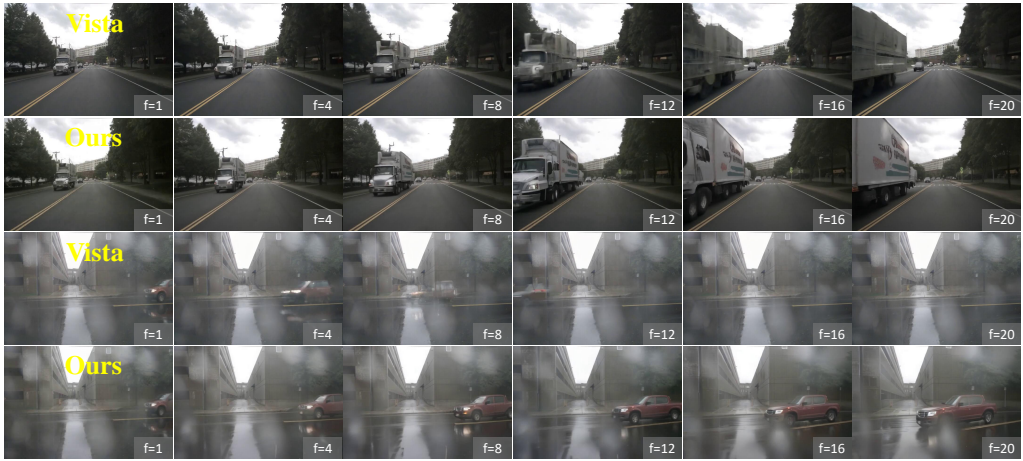}
    \caption{\textbf{Short video prediction comparison.} Compared to our model's prediction with Vista, our model can produce more detailed future frames and generate reasonable and realistic motion.}
    \label{fig:short}
\end{figure*}

\paragraph{Automatic Evaluation}
Following Vista~\cite{gao2024vista}, we conduct the comparison on all samples of the validation set of NuScenes. Tab.~\ref{tab:fvd} presents the comparison of prediction fidelity of the state-of-the-art models. Compared with previous models without using extra driving videos for training, our model outperforms methods using U-Net or DiT such as GenAD and CogVideoX$^\texttt{sft}$ on video fidelity by large gains, and obtains better image fidelity than them. Compared with SOTA models (GenAD, Vista, UniMLVG) using extra driving videos for training, our model also outperforms them on video fidelity. We also report the performance of the reconstruction using CogVideoX VAE, which can be regarded as the FID and FVD lower bounds. The evaluation results show that our models' video prediction quality is closer to FVD's lower bounds. The short video prediction comparison is shown in Fig.~\ref{fig:short}. We adopt more metrics for measuring temporal coherence from VBench~\cite{huang2024vbench}, as shown in Tab.~\ref{tab:vbench}, our model surpasses Vista in dynamic degree and imaging quality by a large margin.

\begin{figure}
    \centering
    \includegraphics[width=\linewidth]{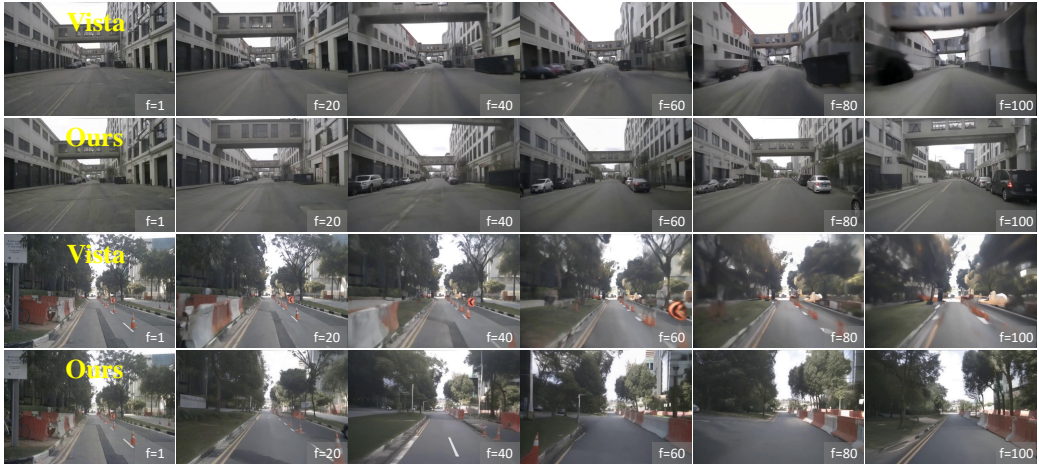}
    \caption{\textbf{Long-term video prediction comparison.} Compared with the state-of-the-art Vista~\cite{gao2024vista} which faces severe error accumulation, our model can predict higher-quality scenarios for the long-term future and generalize to diverse scenarios.}
    \label{fig:longvid}
\end{figure}

\begin{figure}
    \centering
    \includegraphics[width=0.9\linewidth]{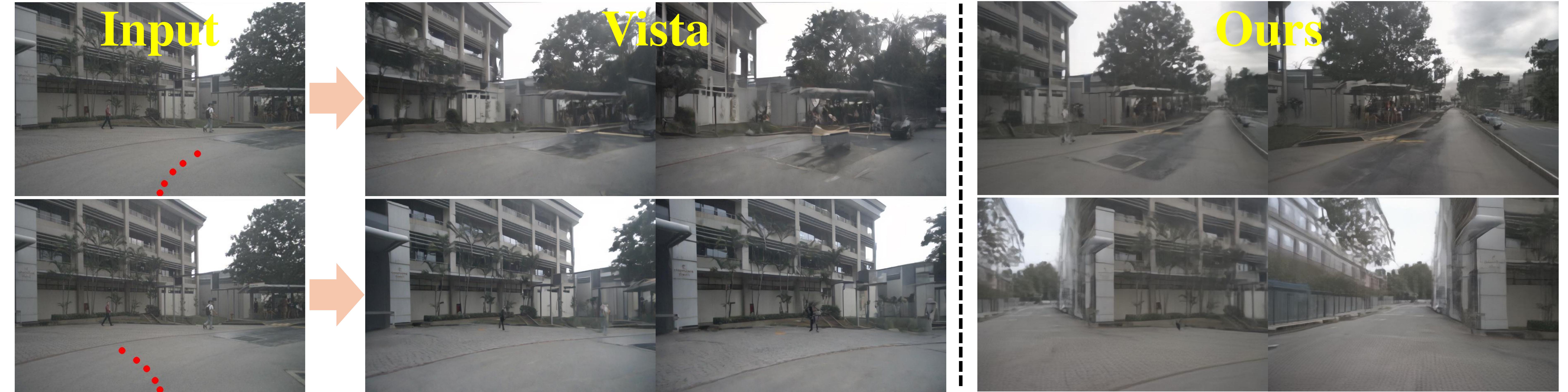}
    \caption{ \textbf{Trajectory controllability comparison.} For different input images and trajectory points, we present the generation results (the 12th and 24th frames) of Vista and our model. Our model not only generates more realistic scenes but also accurately predicts the future under varying trajectories. This highlights the counterfactual reasoning capability of our approach.}
    \label{fig:traj}
\vspace{-20pt}
\end{figure}



\begin{table}[t]
\vspace{-5mm}
\centering
\begin{minipage}{0.35\textwidth}
    \centering
    \caption{\textbf{Evaluation for Visual Quality and Motion Rationality.}}
    \label{tab:human}
\resizebox{0.9\textwidth}{!}{
    \begin{tabular}{l|cc}
    \toprule
    \textbf{Model}  & \textbf{V.Q.↑} & \textbf{M.R.↑} \\
    \midrule
    Vista   & 45.5\%   &  47.5\% \\

    \midrule
    Ours      &    \textbf{54.5\%}       & \textbf{52.5\%}  \\
    \bottomrule
    \end{tabular}
    }
\end{minipage}
\hspace{0.03\textwidth} 
\begin{minipage}{0.5\textwidth}
    \centering
    \caption{\textbf{Trajectory-based evaluation}, using validation based on Vista's annotations.}
    \label{tab:taj}
\resizebox{0.8\textwidth}{!}{
    \begin{tabular}{l|cc}
        \toprule
        \textbf{Model} & \textbf{FID↓} & \textbf{FVD↓} \\
        \midrule
        Vista & \textbf{9.3} & 118.8 \\
        CogVideoX + Vista's feat. & 13.0 & 89.8 \\
        \midrule
        Ours & 12.7 & \textbf{69.6} \\
        \bottomrule
    \end{tabular}
    }
\end{minipage}
\vspace{-10pt}
\end{table}

\begin{table}[t]
\centering
\begin{minipage}{0.35\textwidth}
    \centering
    \caption{\textbf{Evaluation for Trajectory Compliance}, using validation based on Vista's annotations.}
    \label{tab:human-taj}
\resizebox{0.9\textwidth}{!}{
    \begin{tabular}{l|cc}
        \toprule
        \textbf{Model} & Vista & Ours \\
        \midrule
        \textbf{Win Rate↑} & 32.1\% &  \textbf{67.9\%} \\
        \bottomrule
    \end{tabular}
}
\end{minipage}
\hspace{0.03\textwidth} 
\begin{minipage}{0.55\textwidth}
    \centering
    \caption{\textbf{Ablation study}, analyze module effects.}
    \label{tab:abl}
\resizebox{\textwidth}{!}{
    \begin{tabular}{l|cccc}
        \toprule
        \textbf{Method/ Module} & DiT  & Coarse-Fine & Distill. & FVD↓  \\
        \midrule
        CogVideoX$^{\texttt{sft}}$ & \cmark &  & & 347.7 \\
        Decouple learning  & \cmark & \cmark & & 300.3 \\
        Full & \cmark & \cmark  & \cmark & 290.6 \\
        \bottomrule
    \end{tabular}
    }
\end{minipage}
\end{table}

\begin{wraptable}{r}{0.4\textwidth}
\vspace{-5pt}
\centering
\caption{VBench evaluation results.}
\vspace{-8pt}
\label{tab:vbench}
\scalebox{0.8}{
\begin{tabular}{l|ccc}
\toprule
\textbf{Model}  & \textbf{Dynamic↑} & \textbf{Imaging↑} & \textbf{Motion↑} \\
\midrule
Vista   & 74.6\% & 48.9\% & \textbf{99.0\%} \\
Ours      & \textbf{93.7\%} & \textbf{50.3\%} & \textbf{99.0\%} \\
\bottomrule
\end{tabular}
}
\vspace{-10pt}
\end{wraptable}

\paragraph{Human Evaluation} Using automatic metrics does not always align with human preference, so we introduce the human evaluation to asses the visual quality and motion rationality of generated videos following~\cite{gao2024vista}. Because Vista significantly outperforms other baselines, based on the human evaluation results in~\cite{gao2024vista}, in order to save the cost of manual annotation, we only conducted a human evaluation of random side-by-side video selection between our method and Vista. Details can be found in Appendix~\ref{app:eval}. As shown in Tab.~\ref{tab:human}, our method outperforms Vista on visual quality and motion rationality.



\paragraph{Results of Long-term Prediction}
To compare the ability of long-term prediction, we set up comparisons with various video frames including 25, 69, and 113 frames, which correspond to the 1st, 3rd, and 5th rollouts of Vista, respectively. Table.~\ref{tab:long} reports FVD and inference time of baselines and our model. As for inference time, other baselines only support autoregressive rollouts, so the time increases linearly, while our model supports parallel interpolation, which reduces the total inference time significantly. Our model achieves comparable FVD with Vista for short videos, but surpasses all baselines on longer durations. Note CogVideoX-I2V-5B uses DiT as the backbone; hence, the comparison demonstrates our model effectively bridges the gap between general and driving scenes.

\subsection{Results of Controllable Prediction}
\paragraph{Automatic Evaluation}
We set up a fair trajectory-based prediction comparison using Vista's annotations and compare our model with Vista. The results are shown in Tab.~\ref{tab:taj}. Our model outperforms Vista on video quality by a significant margin according to FVD. Using the same pretrained models, our warp-guided method is better than the condition method in Vista, which validates that our model can make higher-quality controllable predictions. Fig.~\ref{fig:traj} presents the visualization comparison of trajectory-based prediction. With the same image input and various trajectories, Vista's results have more distortions and non-compliance with trajectory, while our model can make high-quality controllable predictions.
\begin{wraptable}{r}{0.4\textwidth}
\vspace{-10pt}
\centering
\caption{Frame interpolation evaluation.}
\vspace{-5pt}
\label{tab:framer}
\scalebox{0.8}{
\begin{tabular}{l|cc}
\toprule
\textbf{Interpolation Model} & \textbf{FID↓} & \textbf{FVD↓} \\
\midrule
SEINE~\cite{chen2023seine} & 23.4 & 213.6 \\
FRAMER~\cite{wangframer} & 29.0 & 147.6 \\
Fine-DiT (ours) & \textbf{15.8} & \textbf{117.5} \\
\bottomrule
\end{tabular}
}
\vspace{-20pt}
\end{wraptable}

\paragraph{Human Evaluation} For the specific driving scenario, given the input trajectory, metrics like FID and FVD can not measure the compliance, so here we introduce human evaluation to judge the ability of trajectory compliance. As shown in Tab.~\ref{tab:human-taj}, according to human evaluation, our method has twice the winning rate of Vista.

\subsection{Ablation Study}
\begin{wrapfigure}{r}{0.5\linewidth}
    \centering
    \vspace{-30pt} 
    \includegraphics[width=\linewidth]{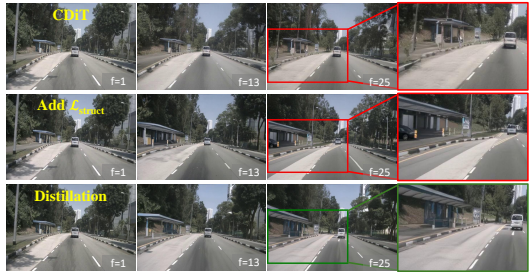}
    \caption{\textbf{Ablation study.} The first row shows CDiT predictions including lots of distortions, the second row adds the structure preservation term, and the last row shows the best.}
    \label{fig:abl}
    \vspace{-26pt} 
\end{wrapfigure}
\paragraph{Quantitative Results}
Tab.~\ref{tab:abl} shows the effect of the three main modules used in our method, including DiT, coarse-to-fine, and distillation. Decouple learning using coarse-to-fine training brings good improvements in the driving scenario, and distillation further improves the temporal consistency. In Tab.~\ref{tab:framer}, our method achieves the best FID and FVD scores compared with two interpolation models~\cite{chen2023seine,wangframer}.
\paragraph{Qualitative Results}
We introduce a structure preservation loss to maintain the quality of each frame. As shown in Fig.~\ref{fig:abl}, comparing the first row and second row, adding the structure preservation term improves the quality and fidelity of each coarse frame, avoiding obvious distortions. We propose the distillation from fine flows to coarse flows to improve the consistency between coarse frames. Comparing the second row and last row in Fig.~\ref{fig:abl}, our distillation improves both the consistency and fidelity of coarse frames.

\vspace{-4mm}
\section{Conclusion}
In this paper, we aim to alleviate the error accumulation problem in the long-term driving world model predictions. Intuitive comparisons suggest that the main challenges are 1) the training-inference gap and 2) the gap between general scenes and driving scenes. To address these, we propose several solutions to build a simple yet effective long-term driving world model. We decouple long video world model learning into large motion learning and bidirectional continuous motion learning, and leverage scalable DiT to solve them. A novel distillation method is proposed to utilize fine-flow priors to guide the coarse-flow to improve the long video consistency, where fine-grained video flows are self-supervised signals for coarse-grained video flows. Extensive experiments on NuScenes demonstrate our model achieving state-of-the-art performance on FVD metrics on various tasks, improving the controllable ability and significantly reducing the inference time for long-term videos.

\bibliographystyle{unsrt}
\bibliography{neurips}

\clearpage
\appendix


\startcontents[chapters]
\setcounter{section}{0}
\printcontents[chapters]{}{1}{}

\section{Details of warp-guided controllable video prediction}
\label{app:detal_warp}
\subsection{Trajectory to camera pose transformation.} Directly using camera poses as condition input is straightforward, but comes with challenges such as construction difficulties and a lack of intuitiveness. Since our model is primarily designed for driving scenarios, simple 2D trajectory points can effectively capture the motion process. Therefore, we follow Vista's~\cite{gao2024vista} trajectory point representation for condition input and apply linear transformations to convert these points into camera poses. Given that Vista provides only limited 2D trajectory points $\{(x'_i,z'_i), 
i< 5 \}$ but requires the generation of a video with $N=25$ frames, so we apply B-spline interpolation $f$ to estimate the 3D trajectory point $Traj(t)=(x_t,0,z_t)=f(t), f: [0, N) \rightarrow \text{span}\{(x'_i, z'_i)|i<5\}$ for each frame. For simplicity, we assume that the camera plane is always perpendicular to the trajectory curve. Then, we can obtain the translation vector $T_t \in \mathbb{R}^{3}$ the camera rotation matrix $R_t \in \mathbb{R}^{3 \times 3}$ at time $t$.

\subsection{Depth-free image warping.} Previous warp-based camera control methods ~\cite{hou2024training} rely on depth information and require more effort. Due to the advanced generative capabilities of the diffusion model, accurate depth information is not necessary. Therefore, we propose an efficient method for image warping that does not require depth.
For an input image $I_0$, we assume that each pixel has the same depth $d$, which allows us to easily back-project the image into 3D space and then project it onto the warped image $\hat{I}_t$ using the camera pose $[R_t,T_t]$. Given the camera intrinsic matrix $K \in \mathbb{R}^{3 \times 3}$, the relationship between $I_0$ and $\hat{I}_t$ can be formulated as:
\begin{equation}
\hat{P}_t = K \cdot \left(R_t \cdot K^{-1} \cdot {P_0} \cdot d + T_t \right)
\end{equation}
where $\hat{P}_t$ and $P_0$ represent the homogeneous coordinates in image $\hat{I}_t$ and $I_0$ respectively. To account for the sparsity and occlusion of the points, we then interpolate the warped image using the Navier-Stokes algorithm ~\cite{bertalmio2001navier}, which is based on fluid mechanics.

\section{Anotations Details}
We use LLaVA-NeXT-Video-7B~\cite{zhang2024llavanextvideo} to annotate each video sequence of 8 keyframes. We obtain annotations for all samples in the training set and validation set of NuScenes. The used prompt is shown in Tab.~\ref{tab:prompt}.
\begin{table}[h]
\centering
\begin{tabular}{|p{0.95\linewidth}|}
\hline
\rowcolor[gray]{0.8}
\textbf{Prompt for MLLM annotations} \\ \hline
\rowcolor[gray]{0.95}
Question: A chat between a curious user and an artificial intelligence assistant. The assistant gives helpful, detailed, and polite answers to the user's questions. \\
\rowcolor[gray]{0.95}
\{DEFAULT\_VIDEO\_TOKEN\} \\
\rowcolor[gray]{0.95}
This is a video of a car driving from a front-view camera. Please answer the following questions based on the video content. Follow the output format below. Answers should be clear, not vague. \\
\rowcolor[gray]{0.95}
\ \ Weather: (e.g., sunny, cloudy, rainy, etc.)\\
\rowcolor[gray]{0.95}
\ \ Time: (e.g., daytime, nighttime, etc.)\\
\rowcolor[gray]{0.95}
\ \ Road environment:\\
\rowcolor[gray]{0.95}
\ \ Critical objects:\\
\rowcolor[gray]{0.95}
ASSISTANT:\\
\rowcolor[gray]{0.95}
Response:\\
\hline
\end{tabular}
\caption{Prompt for MLLM annotation.}
\label{tab:prompt}
\end{table}

\section{Training Details}
\label{app:training}
\paragraph{Coarse Diffusion Transformer.} We initialize the Coarse DiT from CogVideoX-T2V-2B model~\cite{yang2024cogvideox}. The sequence length of training video frames is 13, the resolution is 720$\times$ 480, and the fps is set to 1. We use Diffusers as the training codebase. The Coase DiT is trained on all sampling samples of the training set of NuScenes~\cite{caesar2020nuscenes}, with a total batch size of 8, and trained with 10k steps on 4$\times$A100 GPUs.
\paragraph{Fine Diffusion Transformer.} We also initialize the Fine DiT from CogVideoX-T2V-2B model. The sequence length of training video frames is 13, the resolution is 720$\times$ 480, and the fps is set to 10. The Fine DiT is trained on all continuous frames of the training set of NuScenes, with a total batch size of 8, and trained with 10k steps on 4$\times$A100 GPUs.

\paragraph{Distillation.} After obtaining the well-trained Coarse DiT and Fine DiT, we establish the distillation training. We sample $12\times11+13=145$ frames from a clip for each training sample. We freeze the Fine DiT and train the Coarse DiT using distillation loss for 500 steps with a total batch size of 4 on 4$\times$A100 GPUs.

\section{Evaluation Details}
\label{app:eval}
We evaluate the FID and FVD following Vista~\cite{gao2024vista}, and utilize the implementation from the package: torchmetrics.image.fid.FrechetInceptionDistance.

\paragraph{Human Evaluation}
We conducted a user study to evaluate the visual quality and motion rationality following Vista. We randomly curate 40 scenes from the NuScenes validation set and collect a total of 1600 answers from 20 participants. As for the evaluation of Trajectory compliance, we collected 800 answers in total.

\paragraph{Short Video prediction.} We input the first frame to all models to predict the future 24 frames, where the first frame is from all samples from NuScenes validation set. All 25 frames are used to evaluate the FVD, and only future 24 frames are used to evaluate the FID, which is the same as Vista~\cite{gao2024vista}.
\paragraph{Long Video prediction.} We input the first frame to all models, and rollout multiple times to generate long videos. Different models utilize different rollout times due to different output frames of inference. The first rollout of Vista uses the first input frame, but in subsequent rollouts, it uses the last 3 frames of the previous rollout, so the number of frames Vista generates is: 25, 25+22, 25+22+22, and so on. Each rollout in Vista needs 95s. As for our model with the training sequence of 13, for Coarse DiT, the first rollout generates 13 frames, which cost 35s. Then, between two coarse frames, we can utilize the Fine DiT to generate 11 continuous frames, the maximum frame number is $12*11+13=145$ frames. Because interpolated frames can be processed in parallel, the actual time taken is only the time of one rollout. So for every 145 frames generated, our model only needs 70s. To save the evaluation resources, we chose the first 5 keyframe samples per scene, a total have 750 samples for long video inference. We calculate the FVD scores for 25 frames, 68 frames, and 113 frames respectively. The inference time is tested on a single A100.

\paragraph{Controllable Video prediction.} We use the trajectory annotations from Vista. We utilize the index: [0, 3, 6, 9, 12, 15, 18, 21, 24, 27] to select samples from each scene according to the annotations. So we finally use a total of 1500 samples to evaluate the FVD and FID. We input the first frame into models and get a total of 25 frames for evaluation. 

\paragraph{Rollout Details.} We present the rollout details of compared baselines and our model in Tab.~\ref{tab:roll}. The methods in the first four columns (I2VGen-XL, DynamiCrafter, CogVideoX-I2V-5B, SVD) all use only one frame as the conditional frame during rollout, Vista uses three frames as conditional frames during rollout, while our model can interpolate frames in parallel, so more frames can be generated in parallel during the second rollout.

\paragraph{Distillation Details} We use widely used metrics in VBench [1] to compare the generated videos' quality among different distillation steps. By default, we set the distillation step to 100, which has a lower training cost and better motion smoothness.

\begin{table}[]
    \centering
    \caption{\textbf{Trade-off between distillation steps and model quality.}}
    \label{tab:distill-vbench}
    \begin{tabular}{l|cccc}
    \toprule
    \textbf{Step} / \textbf{Metric} & \textbf{Motion Smoothness} & \textbf{Imaging quality} & \textbf{Aesthetic Quality} &  \textbf{Avg} \\
    \midrule
      50   & 94.6\% & \textbf{67.1\%} & 55.1\%  &  72.4\%  \\
      100  & \textbf{95.0\%} & 64.7\% & 54.9\%  &  71.5\%  \\
      150  & 94.7\% & 65.5\% & 54.8\%  &  71.7\%  \\
      200  & 94.6\% & 64.7\% & \textbf{57.8\% } &  72.4\%  \\
      250  & 94.9\% & 66.2\% & 56.4\%  &  \textbf{72.5\%}  \\
    \bottomrule
    \end{tabular}
    
\end{table}

\begin{table*}[h]
\caption{\textbf{Long-term video prediction rollout details}}
\label{tab:roll}
\centering
\resizebox{\textwidth}{!}{
    \begin{tabular}
    {lcccccc}
    \toprule
     \textbf{Model} & I2VGen-XL & DynamiCrafter& CogVideoX-I2V-5B &   SVD    &   Vista  & Ours  \\

     \textbf{Frames per rollout} & 25  &  16  & 49 & 25  & 25 & 13 \\
     \textbf{Time per rollouts} & 78s  & 88s  & 140s  & 44s  & 95s  & 35s  \\

     \textbf{Total frames using 2 rollouts} & 25+24  & 16+15  & 49+48  & 25+24  & 25+22  & 13+12*11  \\
     \textbf{Total time using 2 rollouts} & 78*2s  & 88*2s  & 140*2s  & 44*2s  & 95*2s  & 35*2s  \\

     \bottomrule

    \end{tabular}
    }
\end{table*}


\end{document}